\documentclass[runningheads]{llncs}

\usepackage{graphicx}
\usepackage{caption}   % must be loaded first
\usepackage{subcaption}
\usepackage{multirow}
\usepackage{booktabs}
\usepackage{gensymb}
\usepackage{amsmath}

\usepackage[hidelinks]{hyperref}

\title{Unified 3D MRI Representations via Sequence-Invariant Contrastive Learning}
\titlerunning{SeqInv: Sequence-Invariant Contrastive Learning for 3D MRI}

\author{
  Liam Chalcroft\inst{1}\thanks{Corresponding author}\and
  Jenny Crinion\inst{2}\and
  Cathy J.~Price\inst{1}\and
  John Ashburner\inst{1}
}
\authorrunning{Chalcroft \textit{et al.}}

\institute{
  Department of Imaging Neuroscience, University College London, UK\\
  \email{l.chalcroft@cs.ucl.ac.uk}
  \and
  Institute of Cognitive Neuroscience, University College London, UK
}

\begin{document}

\maketitle

% Abstract
\begin{abstract}
Self-supervised deep learning has accelerated 2D natural image analysis but remains difficult to translate into 3D MRI, where data are scarce and pre-trained 2D backbones cannot capture volumetric context. We present a \emph{sequence-invariant} self-supervised framework leveraging quantitative MRI (qMRI). By simulating multiple MRI contrasts from a single 3D qMRI scan and enforcing consistent representations across these contrasts, we learn anatomy-centric rather than sequence-specific features. The result is a single 3D encoder that excels across tasks and protocols. Experiments on healthy brain segmentation (IXI), stroke lesion segmentation (ARC), and MRI denoising show significant gains over baseline SSL approaches, especially in low-data settings (up to +8.3\% Dice, +4.2 dB PSNR). It also generalises to unseen sites, supporting scalable clinical use. Code and trained models are publicly available.
\end{abstract}

% Intro
\section{Introduction}
\label{sec:intro}

Deep learning now underpins medical image registration \cite{Balakrishnan_2019} and segmentation \cite{chalcroft2023largekernelattentionefficientrobust}. However, unique challenges arise when working with 3D MRI data, including increased computational demands and the difficulty of applying 2D pre-trained models to volumetric contexts \cite{Ma_2024}. Although large-scale 3D datasets and models \cite{wu2024largescale3dmedicalimage} have recently emerged, fine-tuning them for specific clinical tasks remains non-trivial due to inevitable domain shifts \cite{yang2024textbookremedydomainshifts}.

Self-supervised learning (SSL) offers a promising means of learning robust representations without the need for large labelled datasets. Yet, existing SSL methods often treat each MRI sequence as a separate domain, neglecting the shared anatomical information across contrast variations. In contrast, we leverage the observation that different MRI sequences, despite their unique contrast properties, encode the same underlying anatomy. Embedding MRI physics in SSL is expected to yield representations that ignore sequence contrast yet keep anatomy.

We (i) introduce a physics-driven, sequence-invariant SSL framework, (ii) boost Dice by up to 8.3\% and PSNR by 4.2 dB with only 1\% labels, and (iii) show strong cross-site generalisation.

We provide comprehensive evaluations of the prosed method on three diverse tasks - healthy brain segmentation, stroke lesion segmentation, and image denoising - highlighting the clinical utility of our approach.

Our method addresses key problems in medical imaging by enabling robust feature learning across different sites and sequences, even with limited annotated data. This work takes a step towards developing more generalisable and clinically applicable models. We release all code and backbone weights at \href{https://www.github.com/liamchalcroft/contrast-squared}{github.com/liamchalcroft/contrast-squared}.

% Related Work
\section{Related Work}
\label{sec:related-work}

We briefly review three core areas that underpin this work: contrastive learning, robust representations in 3D medical imaging, and quantitative MRI (qMRI).

\subsection{Contrastive Learning}
Self-supervised learning (SSL) can leverage unlabelled data by creating proxy tasks that encourage useful invariances in learned representations. Techniques include predictive coding, masked image modelling, and contrastive learning. 

Recent contrastive methods such as SimCLR \cite{chen2020simpleframeworkcontrastivelearning} and MoCo \cite{he2020momentumcontrastunsupervisedvisual} learn representations by aligning features from different augmented views, while BYOL \cite{grill2020bootstraplatentnewapproach} and Barlow Twins \cite{zbontar2021barlowtwinsselfsupervisedlearning} reduce the reliance on explicit negative samples or introduce redundancy reduction.

SSL is now routine in medical imaging for using unlabelled data to boost downstream tasks. Adopted methods include contrastive learning \cite{tang2022selfsupervisedpretrainingswintransformers}, masked image modelling \cite{wang2023swinmmmaskedmultiviewswin} and reconstruction-based proxy tasks \cite{Misra2020Self-supervisedRepresentations,Zhou_2021}.

\subsection{Robust Representations in Medical Imaging}
Clinical MRI segmentation tasks face challenges when transferring models to new hospitals or protocols. Public benchmarks often involve a small set of consistent sequences, limiting models to scenarios where training and testing domains match (e.g.\ T1w-only). Real-world deployment must handle diverse sequences and acquisition conditions.

Existing domain adaptation methods typically require multiple unlabelled images or prior knowledge of the target domain \cite{Dorent2023}, which is not always feasible. SynthSeg \cite{Billot2023} addresses this by randomising tissue contrast with synthetic data, with subsequent work showing the transferrability of the learned representations to new tasks. Their success hinges on synthetic data quality, which may miss fine anatomy. Similarly, \cite{Meyer2021} adjust contrast on specific regions in real images, but this approach is restricted to modest in-domain variations rather than full sequence simulation. Meanwhile, \cite{roschewitz2024robustimagerepresentationscounterfactual} demonstrate that generating counterfactual views can boost domain robustness for 2D chest X-ray encoders. We extend these insights to 3D MRI for sequence-invariant representations.

\subsection{Quantitative MRI}
Quantitative MRI (qMRI) acquires per-voxel parameter maps (\emph{e.g.}, $R_1$, $R_2^*$) that govern the signal formation in conventional scans \cite{Weiskopf2021}. These maps facilitate the simulation of numerous synthetic MRI sequences from a single qMRI acquisition \cite{Tanenbaum2017}, improving model robustness under domain shift. For example, synthesised multi-contrast data has led to enhanced results in healthy brain parcellation \cite{borges2021acquisitioninvariant}, improved visualisation and segmentation of subcortical structures through synthetic multi-inversion-time contrasts \cite{hays2025syntheticmultiinversiontimemagnetic}, and better pathology segmentation \cite{chalcroft2024domainagnosticstrokelesionsegmentation}. Other methods rely on MR fingerprinting \cite{Ma2013} to derive similar quantitative maps \cite{Adams2024}, further expanding opportunities for sequence-invariant learning.

% Methods
\section{Methods}
\label{sec:methods}

\begin{figure*}[h]
    \centering
    \subfloat{
        \includegraphics[width=0.95\linewidth]{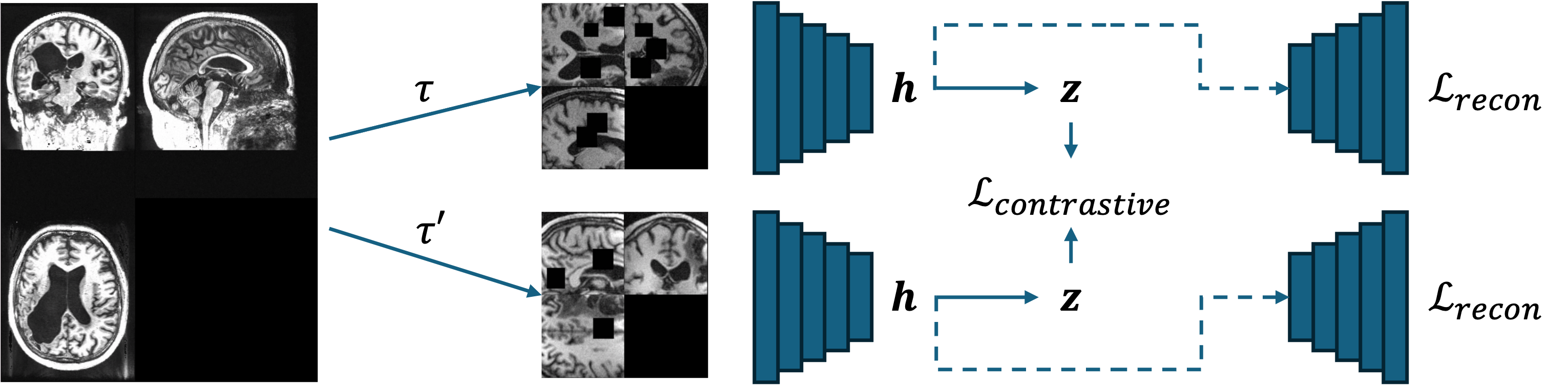}
        \label{fig:ssl-mprage}
    }
    \\
    \subfloat{
        \includegraphics[width=0.95\linewidth]{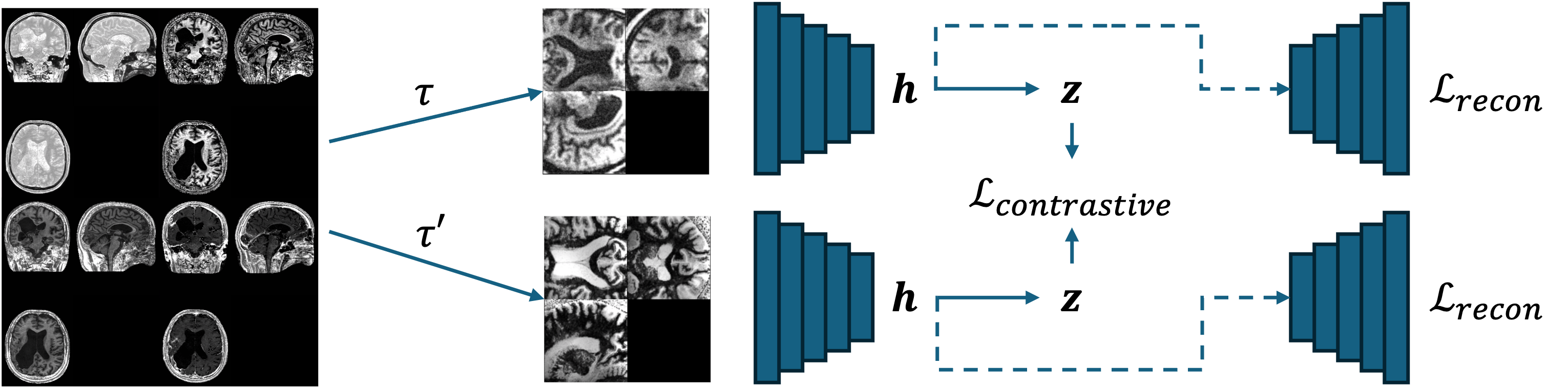}
        \label{fig:ssl-bloch}
    }
    \caption{\textbf{Overview of the proposed SSL approach}. (\ref{fig:ssl-mprage}) \textbf{Baseline}: An MPRAGE volume is augmented into two random views. We extract a feature vector $h$ via the backbone encoder, project it to $z$ for a contrastive loss $\mathcal{L}_{\text{contrastive}}$, and use a decoder to optimise a reconstruction/inpainting loss $\mathcal{L}_{\text{recon}}$. (\ref{fig:ssl-bloch}) \textbf{SeqAug/SeqInv}: We generalise this by simulating multiple scanner sequences from qMRI parameter maps, enabling sequence-invariant learning. In \textbf{SeqAug} both $\tau$, $\tau '$ would produce augmented views of the same sequence, while in \textbf{SeqInv} both views will contain a different sequence.}
    \label{fig:ssl-overview}
\end{figure*}

We propose sequence-invariant SSL for robust 3D MRI representations.
Figure \ref{fig:ssl-overview} shows (i) a contrastive encoder, (ii) a reconstruction decoder, and (iii) a physics engine that simulates multiple sequences from qMRI.

\subsection{Self-Supervised Learning}
\label{sec:methods-ssl}
We adopt SimCLR \cite{chen2020simpleframeworkcontrastivelearning} as our core contrastive framework, though other SSL methods could also be used. Following \cite{tang2022selfsupervisedpretrainingswintransformers}, we incorporate an additional reconstruction branch. Specifically:

\begin{itemize}
    \item \textbf{Contrastive branch:} We create two augmented 3D views of a single input volume. Each view is passed through a shared encoder, producing latent vectors $(z_i, z_j)$. A contrastive loss encourages $z_i$ and $z_j$ to be similar while remaining distinct from other samples in the batch. This step induces a rich feature representation that generalises well across domains.
    \item \textbf{Reconstruction branch:} A lightweight decoder reconstructs the original volume from the latent features after removing artificially added artefacts (\emph{e.g.}, noise, dropout). An $L_{1}$ loss enforces pixel-level fidelity.
\end{itemize}

Spatial augmentations include random crops, rotations, shears and flips. We then apply MRI-specific augmentations such as non-uniform intensity fields, Gibbs artefacts, Rician noise and random cuboid dropout \cite{pathak2016contextencodersfeaturelearning}. In the baseline version, we generate these augmented views from simulated MPRAGE images. Magnetisation-Prepared RApid Gradient Echo (MPRAGE) is a common T1-weighted structural MRI sequence, particularly common in research studies. In our sequence-invariant framework, we instead use parameter maps to simulate diverse MRI sequences (Sec.~\ref{sec:methods-physics}), enabling the encoder to learn anatomy-centric features rather than sequence-specific contrast. We train a model \textbf{SeqAug} that generates two views from a single simulated sequence, and a second model \textbf{SeqInv} that generates the two views from distinct sequence simulations, formally encouraging invariance to choice of MRI sequence. Our baseline model (\textbf{Base}) was pretrained exclusively using synthetic MPRAGE images generated from qMRI parameter maps, ensuring a fair comparison to our proposed methods.

\subsection{Physics-Based Data Synthesis}
\label{sec:methods-physics}
We leverage qMRI maps (PD, $R_1$, $R_{2}^*$, MT) to synthesise multiple MRI contrasts from a single subject. Each voxel’s tissue parameters are passed through forward models approximating various standard MRI sequences (FSE, GRE, FLAIR, MPRAGE). Full signal equations are derived from known relaxation properties (Appendix~\ref{sec:appendix-physics}), and Rician noise is added for realism. By sampling different scanner parameters (\emph{e.g.}, echo time, flip angle), we obtain a range of synthetic images sharing identical anatomical structure but differing in appearance. All simulations use the NITorch library\footnote{https://github.com/balbasty/nitorch}.

\section{Experiments and Results}
\label{sec:results}

We evaluate our sequence-invariant approach on three downstream tasks: healthy brain segmentation, stroke lesion segmentation, and MRI denoising. Following standard practice, we measure segmentation performance using the Dice Similarity Coefficient (DSC) and 95th percentile Hausdorff Distance (HD95), and denoising performance using Peak Signal-to-Noise Ratio (PSNR).

\subsection{Implementation Details and Data Setup}

\paragraph{Pretraining.}
We pre-train three encoders: Base (real MPRAGE only), SeqAug (two views of one simulated sequence) and SeqInv (views from two simulated sequences). Architecture details are given in Appendix \ref{app:architectures}.

All models use NT-Xent \cite{chen2020simpleframeworkcontrastivelearning} (temperature 0.5) plus an equally-weighted $L_{1}$ reconstruction term. The pretraining dataset consists of 51 qMRI volumes (22 healthy, 29 stroke subjects), with sequence simulation performed using Bloch equations for \textbf{SeqAug} and \textbf{SeqInv}.

\paragraph{Downstream Tasks.}
Once pretraining is complete, we freeze the encoder and optimise a U-Net decoder for:
\begin{itemize}
    \item \textbf{Healthy Brain Segmentation:} T1w, T2w, and PDw volumes from the IXI dataset \cite{Robinson2010}, segmented into background, grey matter, white matter, and CSF. We train on $96^3$ patches with affine and intensity augmentations, using a combined Dice + cross-entropy loss. For training, a maximum of 226 subjects are available from the GST site, with 31 reserved for validation and a further 65 for the in-domain test set. For out-of-domain testing, there are 185 and 74 subjects available in the HH and IOP sites respectively.
    \item \textbf{Stroke Lesion Segmentation:} T1w, T2w, and FLAIR from the ARC dataset \cite{Gibson2024TheRepository}. Lesions are often small, so we employ higher class weighting. We use $96^3$ patches and the same augmentations, optimising a combined Dice + cross-entropy loss. The T1w, T2w and FLAIR sequences are distributed in respective train/validation/test splits of (142/20/41), (159/22/47) and (59/8/18).
    \item \textbf{MRI Denoising:} We add synthetic noise ($\sigma=0.2$) to clean IXI scans normalised to a zero mean and unit standard deviation. The network predicts the noise, which is subtracted from the input to produce the denoised image. We evaluate the result via PSNR on the same IXI splits used for healthy segmentation.
\end{itemize}
All models use $96^3$ patches with standard augmentations. Training details including optimization strategy, learning rate schedules, and batch sizes are provided in Appendix \ref{app:architectures}. A new decoder is trained for each task/model.

\subsection{Evaluation Metrics}

\paragraph{Peak Signal-to-Noise Ratio (PSNR)} 
Assesses image quality by comparing the maximum possible signal with the noise. For an image with maximum pixel value $L$, $\text{PSNR} = 20 \cdot \log_{10}\bigl(\tfrac{L}{\sqrt{\text{MSE}}}\bigr)$, where $\text{MSE} = \tfrac{1}{n}\sum_{i=1}^n (y_i - \hat{y}_i)^2$ measures the average error between predicted and ground truth images. While PSNR clearly quantifies denoising improvements, perceptual metrics such as SSIM or LPIPS might provide better insight into human-perceived image quality and will be explored in future analyses.

\paragraph{Dice Similarity Coefficient (DSC)} 
Measures overlap $\text{DSC}(Y,\hat{Y}) = \tfrac{2|Y \cap \hat{Y}|}{|Y| + |\hat{Y}|}$ between a predicted segmentation $\hat{Y}$ and ground truth $Y$. Values range from 0 (no overlap) to 1 (perfect overlap).

\paragraph{95\% Hausdorff Distance (HD95)} 
Reflects boundary accuracy by measuring the 95th percentile of all directed distances between segmentation boundaries. Smaller values indicate better delineation of anatomical edges.

\subsection{Quantitative Results}

\subsubsection{Healthy Brain Segmentation}
Table~\ref{tab:healthy_segmentation_dsc_results} compares DSC scores for T1w, T2w, and PDw images from the IXI dataset, with varying training data proportions (1\%, 10\%, 100\% of the total available data). Our sequence-invariant (\textbf{SeqInv}) model consistently outperforms the baseline (\textbf{Base}), especially in low-data settings and out-of-domain sites (HH, IOP). Meanwhile, the sequence-augmented (\textbf{SeqAug}) model provides moderate gains, particularly on T2w. In the 1\% training data regime, \textbf{Base} performs particularly well on the T2w data; looking at the individual tissue class metrics in Table \ref{tab:healthy_segmentation_dsc_results}, it appears that this is most notable in the White Matter tissue class. We posit that this may be due to the \textbf{SeqAug}/\textbf{SeqInv} models' aversions to learning sequence-specific features preventing them from easily leveraging domain-specific cues such as White Matter in T2-weighted MRI being much darker than any surrounding tissue.

\begin{table}[htbp]
\centering
\caption{Healthy brain tissue segmentation performance using Dice Similarity Coefficient (higher is better). Values show mean ± standard error, with \textbf{bold} and \underline{underlined} indicating best and second-best results. GST represents the training domain.}
\label{tab:healthy_segmentation_dsc_results}
\resizebox{\textwidth}{!}{
\begin{tabular}{lccc|ccc|ccc}
\toprule
 & \multicolumn{3}{c}{\textbf{1\% Training Data}} & \multicolumn{3}{c}{\textbf{10\% Training Data}} & \multicolumn{3}{c}{\textbf{100\% Training Data}} \\
\cmidrule(lr){2-10}
& \textbf{Base} & \textbf{SeqAug} & \textbf{SeqInv} & \textbf{Base} & \textbf{SeqAug} & \textbf{SeqInv} & \textbf{Base} & \textbf{SeqAug} & \textbf{SeqInv} \\
\midrule
\multicolumn{10}{l}{\textbf{In Domain}} \\
\midrule
\textbf{GST [T1w]} & \underline{55.1 ± 0.8} & 38.5 ± 0.9 & \textbf{56.0 ± 0.9} & \textbf{69.3 ± 0.6} & 67.2 ± 0.7 & \underline{67.9 ± 0.7} & \textbf{89.6 ± 0.3} & 84.1 ± 0.6 & \underline{85.5 ± 0.5} \\
\textbf{GST [T2w]} & \textbf{65.4 ± 0.4} & \underline{56.9 ± 0.5} & 47.7 ± 0.8 & \textbf{84.2 ± 0.3} & \underline{79.0 ± 0.3} & 68.6 ± 0.6 & \underline{90.1 ± 0.2} & \textbf{90.5 ± 0.2} & 90.0 ± 0.2 \\
\textbf{GST [PDw]} & 38.1 ± 1.2 & \underline{46.4 ± 1.1} & \textbf{46.6 ± 0.9} & \textbf{74.9 ± 0.6} & \underline{70.8 ± 0.9} & 69.4 ± 0.8 & \textbf{90.1 ± 0.3} & 89.5 ± 0.4 & \underline{90.1 ± 0.3} \\
\midrule
\multicolumn{10}{l}{\textbf{Out of Domain}} \\
\midrule
\textbf{HH [T1w]} & \underline{49.4 ± 0.6} & 33.0 ± 0.6 & \textbf{57.7 ± 0.6} & \textbf{63.0 ± 0.5} & 59.3 ± 0.5 & \underline{61.1 ± 0.6} & \textbf{81.6 ± 0.3} & 75.5 ± 0.5 & \underline{77.4 ± 0.4} \\
\textbf{HH [T2w]} & \textbf{58.6 ± 0.3} & \underline{53.8 ± 0.3} & 46.5 ± 0.3 & \textbf{75.0 ± 0.4} & \underline{72.0 ± 0.3} & 65.6 ± 0.3 & 87.2 ± 0.3 & \textbf{89.7 ± 0.2} & \underline{88.1 ± 0.3} \\
\textbf{HH [PDw]} & 33.8 ± 0.8 & \underline{39.4 ± 0.7} & \textbf{40.3 ± 0.6} & \underline{60.5 ± 0.6} & \textbf{61.5 ± 0.6} & 59.7 ± 0.7 & 82.7 ± 0.4 & \underline{83.1 ± 0.4} & \textbf{85.6 ± 0.4} \\
\textbf{IOP [T1w]} & \underline{50.6 ± 1.3} & 30.7 ± 1.2 & \textbf{54.4 ± 1.0} & \underline{58.3 ± 1.1} & \textbf{60.9 ± 1.2} & 57.4 ± 1.3 & \textbf{79.1 ± 0.9} & 70.7 ± 1.1 & \underline{74.0 ± 0.9} \\
\textbf{IOP [T2w]} & \textbf{58.3 ± 0.6} & \underline{43.8 ± 0.6} & 40.6 ± 0.9 & \textbf{74.7 ± 0.4} & \underline{71.4 ± 0.4} & 63.6 ± 0.7 & 85.1 ± 0.3 & \underline{85.8 ± 0.3} & \textbf{86.1 ± 0.3} \\
\textbf{IOP [PDw]} & 31.1 ± 1.4 & \underline{36.6 ± 1.4} & \textbf{37.6 ± 1.1} & \underline{59.2 ± 0.9} & 55.3 ± 1.1 & \textbf{59.7 ± 0.9} & \underline{76.4 ± 0.7} & 76.3 ± 0.8 & \textbf{77.2 ± 0.7} \\
\bottomrule
\end{tabular}
}
\end{table}

\subsubsection{Stroke Lesion Segmentation}
We next evaluate on the ARC dataset \cite{Gibson2024TheRepository} using both DSC and HD95 (see Table~\ref{tab:stroke_results}). \textbf{SeqInv} achieves the best overall performance on T1w, improving DSC by 0.5 points and reducing HD95 by 5.9 mm compared to the baseline. On T2w, \textbf{SeqAug} reduces HD95 by 22.2 mm, indicating excellent boundary accuracy while maintaining a competitive DSC. For FLAIR, \textbf{SeqInv} provides a further 4.7 mm decrease in HD95, offering improved boundary delineation over the baseline.

\begin{table}[htbp]
\centering
\caption{Stroke lesion segmentation performance using 100\% training data. Values show mean ± standard error, with \textbf{bold} and \underline{underlined} indicating best and second-best results for each metric. DSC (higher is better) and HD95 in mm (lower is better) are shown for each model.}
\label{tab:stroke_results}
\resizebox{0.75\textwidth}{!}{
\begin{tabular}{lcccccc}
\toprule
& \multicolumn{3}{c}{\textbf{DSC}} & \multicolumn{3}{c}{\textbf{HD95 (mm)}} \\
\cmidrule(lr){2-4} \cmidrule(lr){5-7}
& \textbf{Base} & \textbf{SeqAug} & \textbf{SeqInv} & \textbf{Base} & \textbf{SeqAug} & \textbf{SeqInv} \\
\midrule
\textbf{ARC [T1w]} & \underline{78.4 ± 2.0} & 77.3 ± 2.3 & \textbf{78.9 ± 1.9} & \underline{33.2 ± 4.1} & 36.3 ± 4.8 & \textbf{27.3 ± 3.7} \\
\textbf{ARC [T2w]} & 78.7 ± 1.6 & \textbf{80.3 ± 1.4} & \underline{79.4 ± 1.6} & 36.2 ± 3.9 & \textbf{14.0 ± 1.9} & \underline{24.5 ± 3.6} \\
\textbf{ARC [FLAIR]} & 68.4 ± 6.3 & \underline{71.0 ± 5.3} & \textbf{71.1 ± 5.4} & \underline{67.9 ± 4.8} & 68.1 ± 4.3 & \textbf{63.2 ± 3.3} \\
\bottomrule
\end{tabular}
}
\end{table}

\subsubsection{MRI Denoising}
Lastly, we evaluate PSNR on IXI volumes corrupted with synthetic noise (Table~\ref{tab:denoise_psnr_results}). \textbf{SeqInv} achieves notable gains on T1w, boosting PSNR by up to 4.2 dB with only 1\% training data, and these gains persist even at 100\% training data, suggesting robust feature learning. Out-of-domain generalisation is also particularly strong, with \textbf{SeqInv} reaching 21.7 dB on HH T1w compared to 19.3 dB for the baseline. By contrast, \textbf{SeqAug} provides moderate gains, indicating that purely contrast-based augmentation alone cannot match the full sequence-invariant approach. It is notable that the SeqInv model's benefit is much more apparent in this denoising task compared to the previous segmentation tasks. This could be explained by the similarity of image restoration tasks to the objective of contrastive learning to learn invariance to view augmentations. The heavier constraint on invariance due to view-dependent sequences may be better suited to image restoration tasks than discriminative tasks like segmentation and classification.

\begin{table}[htbp]
\centering
\caption{Image denoising performance using Peak Signal-to-Noise Ratio in dB (higher is better). Values show mean ± standard error, with \textbf{bold} and \underline{underlined} indicating best and second-best results. GST represents the training domain.}
\label{tab:denoise_psnr_results}
\resizebox{\textwidth}{!}{
\begin{tabular}{lccc|ccc|ccc}
\toprule
 & \multicolumn{3}{c}{\textbf{1\% Training Data}} & \multicolumn{3}{c}{\textbf{10\% Training Data}} & \multicolumn{3}{c}{\textbf{100\% Training Data}} \\
\cmidrule(lr){2-10}
& \textbf{Base} & \textbf{SeqAug} & \textbf{SeqInv} & \textbf{Base} & \textbf{SeqAug} & \textbf{SeqInv} & \textbf{Base} & \textbf{SeqAug} & \textbf{SeqInv} \\
\midrule
\multicolumn{10}{l}{\textbf{In Domain}} \\
\midrule
\textbf{GST [T1w]} & 14.9 ± 0.0 & \underline{16.2 ± 0.0} & \textbf{19.1 ± 0.0} & 19.0 ± 0.1 & \underline{19.7 ± 0.1} & \textbf{20.3 ± 0.1} & 19.1 ± 0.0 & \underline{20.6 ± 0.0} & \textbf{21.3 ± 0.1} \\
\textbf{GST [T2w]} & 17.2 ± 0.0 & \textbf{17.7 ± 0.0} & \underline{17.3 ± 0.0} & 18.3 ± 0.0 & \underline{18.5 ± 0.1} & \textbf{19.8 ± 0.0} & 18.3 ± 0.0 & \underline{19.4 ± 0.0} & \textbf{20.0 ± 0.0} \\
\textbf{GST [PDw]} & 17.0 ± 0.0 & \underline{18.3 ± 0.0} & \textbf{18.7 ± 0.0} & 18.6 ± 0.0 & \underline{19.3 ± 0.1} & \textbf{20.0 ± 0.1} & 18.7 ± 0.0 & \underline{19.9 ± 0.0} & \textbf{20.6 ± 0.0} \\
\midrule
\multicolumn{10}{l}{\textbf{Out of Domain}} \\
\midrule
\textbf{HH [T1w]} & 15.1 ± 0.0 & \underline{16.5 ± 0.0} & \textbf{19.4 ± 0.0} & 19.1 ± 0.0 & \underline{20.0 ± 0.1} & \textbf{20.1 ± 0.1} & 19.3 ± 0.0 & \underline{21.0 ± 0.0} & \textbf{21.7 ± 0.0} \\
\textbf{HH [T2w]} & \underline{16.5 ± 0.0} & \textbf{16.9 ± 0.0} & 16.4 ± 0.0 & \underline{17.5 ± 0.0} & 15.6 ± 0.1 & \textbf{18.8 ± 0.0} & 17.5 ± 0.0 & \underline{18.5 ± 0.0} & \textbf{18.9 ± 0.0} \\
\textbf{HH [PDw]} & 16.5 ± 0.0 & \textbf{17.8 ± 0.0} & \underline{17.8 ± 0.0} & 18.0 ± 0.0 & \textbf{19.2 ± 0.0} & \underline{18.9 ± 0.1} & 18.2 ± 0.0 & \underline{19.3 ± 0.0} & \textbf{19.9 ± 0.0} \\
\textbf{IOP [T1w]} & 14.7 ± 0.0 & \underline{16.7 ± 0.0} & \textbf{18.9 ± 0.0} & 18.4 ± 0.0 & \textbf{19.9 ± 0.0} & \underline{18.5 ± 0.1} & 18.8 ± 0.0 & \underline{20.3 ± 0.0} & \textbf{21.0 ± 0.0} \\
\textbf{IOP [T2w]} & \underline{17.1 ± 0.0} & \textbf{17.6 ± 0.0} & 17.0 ± 0.0 & 17.9 ± 0.0 & \underline{18.8 ± 0.0} & \textbf{19.6 ± 0.0} & 18.0 ± 0.0 & \underline{19.2 ± 0.0} & \textbf{19.7 ± 0.0} \\
\textbf{IOP [PDw]} & 16.9 ± 0.0 & \underline{18.3 ± 0.0} & \textbf{18.8 ± 0.0} & 18.5 ± 0.0 & \underline{19.8 ± 0.0} & \textbf{20.0 ± 0.0} & 18.6 ± 0.0 & \underline{19.8 ± 0.0} & \textbf{20.5 ± 0.0} \\
\bottomrule
\end{tabular}
}
\end{table}

% Discussion
\section{Discussion}
\label{sec:discussion}

Our results show that sequence-invariant self-supervised learning substantially improves model robustness and generalisation across diverse MRI sequences and acquisition sites. In particular, it enables effective feature learning even with minimal labelled data, suggesting that the method captures fundamental anatomical cues independent of sequence-specific contrast.

\subsection{Key Findings}
We highlight three key aspects of our method’s performance. First, even when trained on as little as 1\% of the data, it achieves up to +4.2 dB PSNR in denoising and +8.3 DSC points in segmentation, underscoring its robust representation capabilities. Second, the model generalises well across T1w, T2w, and PDw, showing particularly strong results on T1w while leaving some gaps on the other sequences. Finally, it excels at out-of-domain adaptation, often surpassing baseline models more in unseen sites than in the original training domain, illustrating its effectiveness for cross-site generalisation.

\subsection{Limitations}
Our approach faces several limitations. Full-resolution 3D training is costly, so batch size - and thus negative pairs - is limited. Second, we rely on a CNN backbone, which may not capture long-range dependencies as effectively as vision transformers or other recent architectures. Third, while cross-sequence invariance bolsters model robustness, certain sequence-specific gaps - particularly on T2w images - highlight the need for further improvements. Further, qMRI inherently is unable to generate modalities such as SWI, DWI or CT, and therefore may still be liable to domain shifts in the presence of such modalities. A notable limitation was pretraining on only 51 subjects, which is relatively small for SSL frameworks. Scaling pretraining to larger qMRI datasets or synthetically derived qMRI maps from large databases such as the UK Biobank could further enhance representation robustness.

\subsection{Future Directions}
Future work will test ViT encoders, larger qMRI datasets and alternative SSL objectives such as VICReg and DINO. We also expect multi-view contrastive learning and decoder pretraining \cite{asiedu2022decoderdenoisingpretrainingsemantic} to be valuable directions of future work. By leveraging large public datasets of structural MRI, it may be possible to use existing methods for estimating qMRI such as \cite{Borges2024} to generate a large, synthetic dataset to benefit from the scaling effects of self-supervised pre-training.

\subsection{Conclusion}
Sequence-invariant self-supervised learning offers a promising route towards more robust, generalisable medical image analysis. By using physics-informed contrast simulation and contrastive training, we can exploit the shared anatomy across varied MRI sequences and sites. Although challenges remain - especially around computational cost and data availability - our results illustrate the potential for significant gains in low-data scenarios and out-of-domain adaptation. We believe this framework provides a stepping stone toward truly cross-domain, clinically deployable deep learning models in medical imaging.

\subsubsection*{Acknowledgements.}
LC is supported by the EPSRC CDT in Intelligent, Integrated Imaging in
Healthcare (EP/S021930/1) and by the Wellcome Trust (203147/Z/16/Z and
205103/Z/16/Z).  NVIDIA donated the RTX A6000 48 GB GPU used in this research.

\bibliography{main}

\clearpage
\appendix
\section{Physics-based Signal Equations}
\label{sec:appendix-physics}

For each voxel we assume proton density (PD), longitudinal relaxation rate
$R_{1}$, transverse relaxation rate $R_{2}$ ($R_{2}^{*}$ for GRE) and,
optionally, magnetisation transfer (MT).  The receive field is denoted $B_{1}$
and the sequence-specific timing parameters are the repetition time $T_{R}$,
echo time $T_{E}$, inversion time $T_{I}$, excitation spacing $T_{X}$, delay
$T_{D}$ and excitation count $n$.

\begin{table}[h]
\centering
\caption{Forward signal models used for sequence synthesis.}
\label{tab:signals}
\begin{tabular}{@{}lp{0.78\linewidth}@{}}
\toprule
Sequence & Signal equation $S = f(\cdot)$ \\ \midrule
Fast Spin-Echo (FSE) &
  $\text{PD}\,B_{1}\bigl(1 - e^{-R_{1}T_{R}}\bigr)\,e^{-R_{2}T_{E}}$ \\[2pt]
Gradient-Echo (GRE) &
  $\text{PD}\,B_{1}\sin\!\alpha\,(1-\text{MT})\,
   \dfrac{1 - e^{-R_{1}T_{R}}}{1 - \cos\!\alpha\,(1-\text{MT})e^{-R_{1}T_{R}}}\,
   e^{-R_{2}^{*}T_{E}}$ \\[4pt]
FLAIR &
  $\text{PD}\,B_{1}e^{-R_{2}T_{E}}\bigl(1 - 2e^{-R_{1}T_{I}}
   + e^{-R_{1}T_{R}}\bigr)$ \\[2pt]
MPRAGE &
  $\text{PD}\,B_{1}\!\Bigl|\sin\!\alpha
   \frac{1 - e^{-R_{1}T_{R}}}{1 - \cos\!\alpha\,e^{-R_{1}T_{R}}}
   \bigl[1 - (\cos\!\alpha\,e^{-T_{X}R_{1}})^{n}\bigr]e^{-T_{D}R_{1}}
   + 1 - e^{-T_{D}R_{1}}\Bigr|$ \\ \bottomrule
\end{tabular}
\end{table}

\subsection*{Noise Simulation}
Rician corruption is applied on-the-fly:
\[
S_{\text{noisy}}
     = \sqrt{\bigl(S_{\text{MRI}} + \epsilon_{r}\bigr)^{2}
              + \epsilon_{i}^{2}},\quad
\epsilon_{r},\epsilon_{i} \sim \mathcal N(0,\sigma^{2})
\]
as in \cite{Gudbjartsson1995}.  All signal synthesis relies on
NiTorch\footnote{\url{https://github.com/balbasty/nitorch}}.

\subsection*{Acquisition Parameter Sampling}

\begin{table}[h]
\centering
\caption{Sampling ranges for synthetic sequence generation
($\log U$ indicates sampling uniformly in log-space).}
\label{tab:params}
\begin{tabular}{@{}llll@{}}
\toprule
Sequence & $T_{E}$ [s] & $T_{R}$ [s] & Additional parameters \\ \midrule
FLAIR  & $\log U(0.02,0.10)$ & $\log U(0.001,5)$ & $T_{I}\sim\log U(0.001,3)$ \\
FSE    & $\log U(0.001,3)$   & $\log U(0.001,3)$ & – \\ 
MPRAGE & $U(0.002,0.004)$    & $N(23,2.3)$
       & $T_{I}\!\sim\!U(0.6,0.9)$,\;
         $T_{X}\!\sim\!U(0.004,0.008)$,\;
         $\alpha\!\sim\!U(5^{\circ},12^{\circ})$ \\[2pt]
GRE    & $\log U(0.002,0.08)$ & $\log U(0.005,5)$
       & $\alpha\!\sim\!U(5^{\circ},50^{\circ})$ \\ \bottomrule
\end{tabular}
\end{table}

All samples are clamped to physically plausible values (negative draws are
reflected).

\section{Model Architectures}
\label{app:architectures}

\subsection*{Pre-training Architecture}

The pre-training setup consists of three components, described in Table \ref{tab:pretrain}. NT-Xent projection head is 

\begin{table}[h]
\centering
\caption{Network modules used during self-supervised pre-training.}
\label{tab:pretrain}
\begin{tabular}{@{}lllll@{}}
\toprule
Module & Layers / blocks & Kernel & Output dims & Notes \\ \midrule
CNN encoder &
  5 conv-blocks & $3^{3}$ &
  64 $\rightarrow$ 768 &
  instance-norm, GELU, dropout 0.2 \\
Projector &
  MLP(768 → 512 → 128) & – & – &
  NT-Xent projection head \\
Reconstructor &
  4 transposed conv & $2^{3}$ &
  768 $\rightarrow$ 48 &
  L1 reconstruction branch \\ \bottomrule
\end{tabular}
\end{table}

\subsection*{Downstream Task Architectures}
For the denoising task, we use a U-Net architecture that incorporates the pre-trained encoder:

\begin{itemize}
    \item \textbf{CNN U-Net}:
        \begin{itemize}
            \item Input: 3D volume with 1 channel
            \item Encoder: Pre-trained CNN encoder (frozen)
            \item Feature dimensions: (768, 512, 256, 128, 64, 32)
            \item Instance normalization throughout
            \item GELU activation functions
            \item Dropout rate: 0.2
            \item Upsampling: Transposed convolutions
            \item Output: 1 channel (predicted noise)
        \end{itemize}
\end{itemize}

\subsection{Training Details}

The models were trained with the following specifications:

\begin{itemize}
    \item Optimizer: AdamW with gradient clipping at 12.0
    \item Learning rate schedule: $(1 - \frac{epoch}{max\_epochs})^{0.9}$
    \item Loss functions:
        \begin{itemize}
            \item Pre-training: NT-Xent loss + L1 reconstruction loss
            \item Denoising: Mean Squared Error (MSE)
            \item Segmentation: Dice + Cross-Entropy
        \end{itemize}
    \item Patch size: 96×96×96
    \item Mixed precision training
    \item Batch size: 
        \begin{itemize}
            \item Pre-training: 8
            \item Downstream tasks: 2
        \end{itemize}
\end{itemize}

During downstream task training, the pre-trained encoder weights were frozen while the decoder weights were trained from scratch, as evidenced by the weight loading and gradient freezing in the training code.

\end{document}